\documentclass{IEEEtran}
\usepackage{graphicx,amsmath,amssymb}
\usepackage{csquotes}
\usepackage{flushend}

\newtheorem{definition}{Definition}

\makeatletter

\begin{document}
\title{Machine Education: Designing semantically ordered and ontologically guided modular neural networks}

\author{\IEEEauthorblockN{Hussein A. Abbass, Sondoss Elsawah}
\IEEEauthorblockA{\textit{School of Engineering \& IT} \\
\textit{University of New South Wales}\\
Canberra, Australia\\
\{s.elsawah,h.abbass\}@adfa.edu.au}
\and
\IEEEauthorblockN{Eleni Petraki}
\IEEEauthorblockA{\textit{Faculty of Education} \\
\textit{University of Canberra}\\
Canberra, Australia\\
Eleni.Petraki@Canberra.edu.au}
\and
\IEEEauthorblockN{Robert Hunjet}
\IEEEauthorblockA{\textit{Land Division} \\
\textit{Defence Science and Technology}\\
Adelaide, Australia\\
robert.hunjet@dst.defence.gov.au}
}

\maketitle

\begin{abstract}
The literature on machine teaching, machine education, and curriculum design for machines is in its infancy with sparse papers on the topic primarily focusing on data and model engineering factors to improve machine learning. In this paper, we first discuss selected attempts to date on machine teaching and education. We then bring theories and methodologies together from human education to structure and mathematically define the core problems in lesson design for machine education and the modelling approaches required to support the steps for machine education. Last, but not least, we offer an ontology-based methodology to guide the development of lesson plans to produce transparent and explainable modular learning machines, including neural networks. 
\end{abstract}


\section{Introduction}\label{intro}

\IEEEPARstart{W}{ith} recent advances in deep learning and a new-found abundance of data on massive scales, researchers have started to feel the need to follow structured methodological approaches to teaching machines in similar ways humans follow structured teaching approaches when teaching other humans, leading to the fields of machine teaching~\cite{mei2015using}, curriculum design for machines~\cite{peng2018curriculum}, and machine education~\cite{Leu2017NIPS}. 

A common aspect of this domain's literature  is that it is driven by machine learners alone with a focus on data engineering and model engineering to improve machine learning. As such, except for a couple of recent attempts~\cite{zhu2018overview,clayton2019machine,alex2019machine}, the literature on machine teaching and education has not taken inspiration from the more mature  literature on human teaching and education which houses a wealth of pedagogical frameworks, philosophies and theories that could benefit the machine education field.

In this paper, we will present an initial approach  to connect threads from the human education literature to machine education. The remainder of this paper is structured as follows. Section~\ref{Section2} acts as a short background section on current work on machine teaching and education. Section~\ref{Section3}  presents key concepts in pedagogy and curriculum design methodologies in human education, followed by the first attempt to structure machine education in a manner that leverages the significant literature on human education, thereby addressing significant methodological gaps in the machine education literature. We then conclude the paper with a discussion on open challenges in the field.

\section{Machine Teaching and Education}\label{Section2}

The emerging literature of machine-teaching can be traced back to Elman\textquoteright s work in 1993~\cite{elman1993}, which promoted the need to \textquoteleft start small \textquoteright \ and structure learning experiences from simple to more complex. Elman systematically increased the memory of the learner while fixing the complexity of the input to the learner, and was able to teach the machine to learn a \textquoteleft semi-realistic artificial language\textquoteright \  with  increasing levels of competencies. These ideas have anchored the roots of machine-teaching in the literature, which started to emerge as a field only in the last few years. However, similar to Elman, the work is mostly driven by experts in machine learning with intimate knowledge of algorithms, simple tasks, and, in most cases, exiguous grounds in teaching methodologies. Moreover, in the field of evolutionary-learning, other than limited recent attempts~\cite{clayton2019machine}, the research area is in its infancy. To the best of our knowledge, attempts in the evolutionary neural network literature are rare. 

Clayton and Abbass~\cite{clayton2019machine} designed a curriculum suitable for reinforcement learning used within an autonomous agent in shepherding tasks. They extended Dick and Carey\textquoteright s Model for Systematic Instructional Design (SID)~\cite{dick2009systematic,dijkstra2013instructional}, which offers the following ten steps: identification of an instructional goal, conducting instructional analysis, identification of entry behaviours and characteristics, write-up of performance objectives, development of criterion-referenced assessments, development of instructional strategies, development of and/or select instruction, development of and carrying out of formative evaluations, revising the instruction, and conducting summative evaluation.

\begin{figure*}[thp]
\centering
\includegraphics[width=0.95\textwidth]{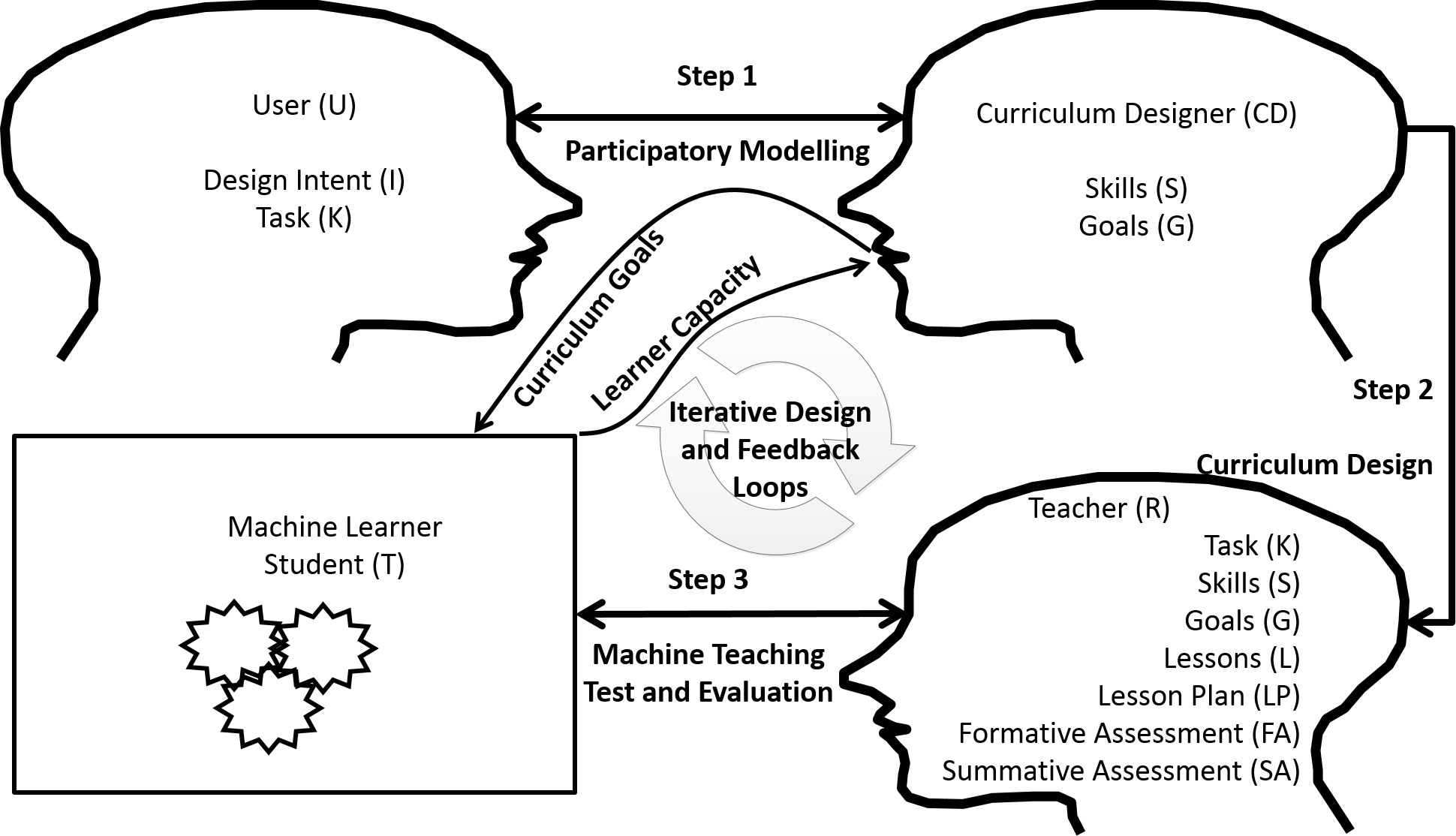}
\caption{The methodological framework for machine education.}
\label{TheoreticalFramework} 
\end{figure*}

Gee and Abbass~\cite{alex2019machine} offered a complete methodology to teach machines by extending George\textquoteright s classical curriculum design~\cite{george2009} with task analysis to suit machine education. They used a supervised learning approach to the shepherding problem to demonstrate the methodology and were successful in learning primitive shepherding behaviours.

Zhu et al.~\cite{zhu2018overview} presented a preliminary high-level study to ground machine teaching in some concepts in human education. They introduced eight dimensions to characterize the different threads of research that have either taken place or could take place in the area of machine teaching: Human vs. machine; Teaching signal; Batch vs. sequential teaching; Model-based vs. model-free; Student awareness; One vs. many; Angelic vs. adversarial; and Theoretical vs. empirical.

Aside from the above, all other research on machine teaching in the literature relies on classic machine learning and deep learning, merely focusing on task decomposition and data engineering.

\section{Machine Teaching and Education}\label{Section3}

The primary challenge in transforming concepts from the human education literature to machine education is the differences between current state-of-the-art in machine cognition and human cognition. Humans can understand and act on ambiguous language sitting at a variety of levels of abstraction. Human teachers follow pedagogical theories that allow them a great deal of freedom in operationalizing the concepts and the fluidity to adopt these concepts differently based on a learner\textquoteright s needs. Machines require a more formal representation of these concepts in the form of mathematical models. To ensure the trustworthiness of the machine learners produced through machine education, assessments, for example, need to be captured in formal models that are mathematically and logically verifiable. At this stage, we need to transform the above concepts mathematically adjust them for a machine context. We will start with the following problem definition, where we will distinguish between machine education, which is concerned with the design and implementation of the process, and machine teaching, which is concerned with the delivery of materials to the machine.

Given a user, $U$, wishing to design an Artificial Intelligence (AI) student, $T$, to fulfil $U$\textquoteright s design intent, $I$, by performing task, $K$, a curriculum designer, $CD$, aims to transform $I$ into a set of concrete learning skills, $S$, and a set of goals, $G$, that concretely measure that $T$ developed $S$ successfully. Given $S$ and $G$, $CD$ needs then to define the set of lessons, $L$, lesson plan, $PL$, a set of formative assessments, $FA$, and a set of summative assessments, $SA$, to teach $S$ to $T$ such that all goals in $G$ are achieved as measured by the formative and summative assessments, $FA$ and $SA$, respectively. If a syllabus, $Y$, is defined by $Y=(L,LP,FA,SA)$, a learner-centric curriculum is defined by the tuple $C=(T,K,S,G,Y)$. Machine education is to design $C$ and select the teacher, $R$, while machine teaching is to select and implement $Y$ such that, $R$ delivers $Y$ to $T$, $T$ develops skills $S$ to perform $K$, so that the goals $G$ are met as measured by $FA$ and $SA$.

We define three broad steps representing the three challenges that need to be overcome to move from the need analysis for curriculum design all the way to a functional syllabus that could be used to teach machine learners. These three steps are presented in the conceptual diagram in Figure~\ref{TheoreticalFramework} and are discussed below: 

\begin{enumerate}

\item Our first step in designing theoretical constructs for machine teaching is to identify an appropriate modelling paradigm to operationalize the human education concepts presented in the previous section and transform them into objective and quantitative models. Formally, this step could be represented as follows:

\[ Step1: (I \wedge K) \rightarrow (S \wedge G) \]

This step will be addressed in the next subsection, which will introduce participatory modelling and the use of system dynamics as a modelling framework suited for representing the complexity that could arise from complex problems in machine learning. The primary aim of these participatory modelling approaches is to transform user requirements in terms of what the user expects the machine learner to learn (need--analysis in curriculum design) into formal specifications of the learning outcomes, formulation of objectives, and models for testing and evaluation of the educational journey of a machine learner.

\item The second step is the design of syllabus, and in particular the decomposition of the overall curriculum into lessons. The following sub-section will focus more on the syllabus and will present the concept of scaffolding as a means of decomposing a learning task into chunks. We will then require that each chunk is associated with appropriate semantics, such that the orders of chunks follow appropriate semantic orders that allow the overall curriculum to be semantically ordered and fit within an appropriate ontological framework.

\[ Step2: (S \wedge G) \rightarrow Y \]

\item The third step focuses on the curriculum design itself. This step is not discussed in this paper is and the context of our future work.

\[ Step3: (S \wedge G \wedge Y) \rightarrow C \]
\end{enumerate}

\subsection{Participatory and system dynamics modelling}

The aim of this step is to transform a user intent and tasks into precise skills and goals; that is, $(I \bigwedge K) \rightarrow (S \bigwedge G)$. Participatory modelling is a purposeful learning process for action that engages the implicit and explicit knowledge of problem actors to create formal and shared representations of the problem of interest~\cite{jordan2018twelve}. The key distinction of participatory modelings from other participatory and actionable science research approaches is its unique focus on the utilization of different types of models (i.e. conceptual and quantitative) and modelling approaches (e.g. system dynamics) to structure the learning process, organize and integrate knowledge (i.e. sources and forms), and package the produced knowledge into useful artefacts. This modelling focus offers a clear stepwise methodological framework to capture an actor's mental model (i.e. perceptions, knowledge, requirements) and transform them into operational outcomes~\cite{voinov2018tools}. 

There exist several typologies for characterising the expected outcomes of employing participatory modelling~\cite{gray2018purpose}. Broadly speaking, operational outcomes can be categorised into: cognitive (e.g. increase in the level of actor\textquoteright s knowledge or skills), social (e.g. establishing formal rules to organize actors\textquoteright \  interactions), and normative outcomes (e.g. development of methods for decision analysis). Participatory modelling has been applied in a variety of domains achieved a wide range of purposes, including decision making~\cite{falconi2017interdisciplinary}, deliberation and negotiation~\cite{barnhart2018embedding}, futuristic system design~\cite{duespohl2016causal}, and to much less extent support formal education and curriculum design. Various participatory modelling frameworks exist that propose specific process steps and combinations of qualitative and quantitative modelling methods~\cite{voinov2018tools}. Major differences among existing participatory modelling frameworks relate to the modelling methods included, such as agent-based modelling (see for example~\cite{elsawah2015methodology}) or system dynamics modelling (see for example~\cite{elsawah2017overview}. Consideration of the context in the design of participatory modelling processes is a critical factor~\cite{elsawah2017overview}.

Regardless of the detailed differences stemming from the choice of the modelling framework, a typical participatory modelling approach proceeds through iterations of the following mechanisms~\cite{voinov2010modelling}: problem scoping and context setting, process design, model development, and feedback, monitoring and evaluation.  
 
Based on systems thinking and control theory, system dynamics encompasses a range of methods including: conceptual and graphical models (e.g. influence diagrams, stock and flow diagrams), quantitative models (e.g. simulation, games~\cite{elsawah2017empirical}), and participatory methods (e.g. groups model building~\cite{scott2016recent}). At the core of the conceptual basis of system dynamics, learning plays a central role in shaping the development of system dynamics methods and praxis. Learning, about complex problems and the mental models that individuals and groups have about these problems, is recognised as a key rationale and outcome for using system dynamics methods~\cite{sterman1994learning}. A typical system dynamics process progresses through the following steps~\cite{sterman2000business}: (1) problem scoping, (2) conceptual modelling, (3) model formulation and implementation, (4) model use.

\begin{figure*}[htb] 
\centering
\includegraphics[width=0.95\textwidth]{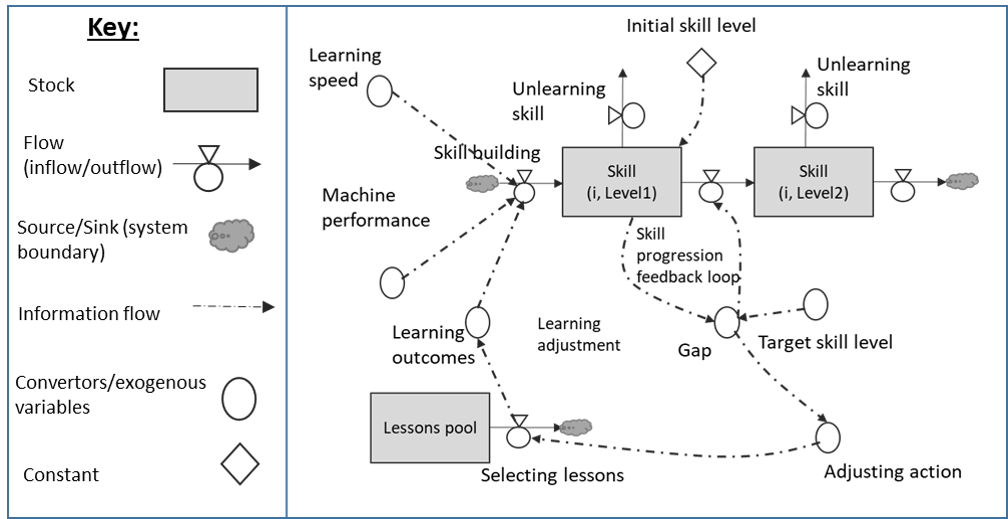}
\caption{High level stock and flow diagram of the progression between skill levels.}\label{SystemDynamic}
\end{figure*}

The process of operationalizing human education concepts into machine education is mainly concerned with addressing the fundamental question of \textquoteleft how can we teach machines effectively\textquoteright ? In other words, what is the relative effectiveness of different pedagogical theories, strategies, tactics, syllabus, and the overall curriculum design in producing the desirable learning outcomes? This is a question that can be addressed through theoretical arguments, but equally through experiments based on simulation of system dynamics models. 
Conceptual system dynamics techniques (such as causal loop diagrams and stocks and flows diagrams) capture the implicit and explicit assumptions about the learning process, with particular focus on interdependency and feedback interactions that shape the learning process and determine its outcomes. Data to support the development and validation of these assumptions, referred to as dynamic hypotheses, are gathered through multiple sources, such as expert opinions, literature surveys, and empirical data~\cite{sterman2000business}. The transformation of the conceptual model into a quantitative system dynamics model allows for testing the effectiveness of different learning theories and implementation scenarios. System dynamics takes the view that any type of dynamic abilities (e.g. skills, knowledge) can be represented as a stock (a rectangle), or a set of related stocks (See Figure~\ref{SystemDynamic} for an example), that accumulates or depletes over time in effect of changes in inflow (e.g. knowledge acquisition, skill building) and outflow (e.g. unlearning skills).

The figure depicts a simplified model of the curriculum 
process, where a gap between the current competency of the machine and its target skill level has been identified. The speed of the learner along with learning outcomes drawn from a lesson pool form the basis upon which skills are built. As skills continue to flow in the mental model of the learner, the first level of skill starts to increase. When the target level of skill is reached, the learner is ready and the skills start to build up in the next level. The control loop continues through a set of adjusting actions that control the draw of lessons from the lesson pool based on observations arising from how the learner is meeting a target skill level. This stock-and-flow diagram has multiple uses. First, it forms the basis for the controller of the machine teaching process that decides when and how much certain data flow to the machine for training. Second, it is a diagnostic tool that helps to identify when the targets for certain skills are not met. Third, it could communicate to, and guide, the teacher (human or machine) in the machine teaching process. The next sub-section will zoom in on stock rectangle labelled as ``lessons pool" to look into how the overall syllabus gets decomposed into lessons.

\subsection{Scaffolding and Syllabus Design}

The aim of this step is to transform the skills and goals from the previous steps into a syllabus; that is, $(S \wedge G) \rightarrow Y$. Recalling that a syllabus is made of lessons, lessons plan, formative, and summative assessment; that is, $Y=(L,LP,FA,SA)$. A key challenge in this step is to transform the skills and goals into lessons and decomposing these lessons in a way that the machine could be taught.

Machine teaching attempts to learn the data set that generated a specific model. This process is a form of inverting a model by projecting it back to the feature space and the data sample that was used to parameterise the model. Equally, if we have a learning outcome that needs to be achieved, machine teaching will generate the learning lessons necessary to reach this learning outcome. For artificially supervised learners, this step is about the generation of the experience, observations, or training samples that are needed to lead to a specific model.

In this section, we propose Incremental Semantic Machine Teaching (ISMT), where the objective is to transform the discovery of the sample to a learning process, whereby the learner starts with a random or targeted sample (called a chunk) then learns to add chunks as necessary to move the learner as close as possible to achieving the overall learning outcomes. 
In this incremental process, the choice of the sample is guided by the semantics to be encoded within the intermediate models.

The concept of scaffolding~\cite{hammond2005putting} in human education involves the decomposition of the curriculum into chunks and the sequencing of these chunks in the learning process to incrementally evolve the learner\textquoteright s knowledge. Hammond and Gibbons~\cite{hammond2005putting} proposed scaffolding at the macro \enquote{designed-in} level. They decomposed the curriculum design process into seven stages: determining student\textquoteright s experience and prior learning, selection of tasks, sequencing of tasks, participant structures, semiotic systems, mediational texts, and metalinguistic and metacognition awareness. We will focus on the formalism for the first four stages and postpone the latter stages to our future work.

Define a target function $\vec{y} = f(\vec{w},\vec{x})$, where $\vec{w}$ are the weights parameterizing the function. The Machine Learning (ML) problem can be defined as:

\begin{definition}
[\bf{ML Problem}] Given a training set $\Gamma=(\vec{x},\vec{y})$, find $\vec{w}$ such that $\hat{\vec{y}} = f(\vec{w} \otimes \vec{x})$ and the loss function, $Err(\hat{\vec{y}},\vec{y})$, is minimum.
\end{definition}

In the above definition, $\vec{x}$ represents the inputs (an $m \times n$ matrix of $m$ observations and $n$, preferably  independent, variables) and $\vec{y}$ represents the true output vector. Together, they form the sample, $\Gamma$, used for building  model $\hat{\vec{y}}$. The error function $Err(\hat{\vec{y}},\vec{y})$ can be arbitrarily chosen to be any type of loss function.

The Machine Teaching (MT) problem is the dual of the ML problem above, where the problem is to find an estimate of the sample $\hat{\Gamma}$, that if it was used in place of $\Gamma$, it would generate an equivalent model $\hat{\vec{y}}$. The problem can be defined formally as follows:

\begin{definition}
[\bf{MT Problem}] Given a model $\hat{\vec{y}} = f(\vec{w} \otimes \vec{x})$ and its parameters $\vec{w}$, find $\hat{\Gamma}$ such that the loss function, $Err(\hat{\vec{y}},\vec{y})$, is minimum.
\end{definition}

One of the greatest statisticians of his time wrote: \textquote[p278, {\cite{fisher1922mathematical}}]{\textit{$\dots$, the object of statistical methods is the reduction of data. A quantity of data, by which usually by its mere bulk is incapable of entering the mind, is to be replaced by relatively few quantities which shall adequately represent the whole, of which, in other words, shall contain as much as possible, ideally the whole, of the relevant information contained in the original data}}.  
 
Even though ML attempts to find this reduced form by describing a model of the data, so often some form of parsimonious pressures are needed to further ensure that the model is the shortest description possible~\cite{barron1998minimum}. Parsimonious pressures in fields such as neural networks are achieved through a regularization term that gets added to the loss function.

Similar to ML where the parsimonious pressure is to find the shortest model, the parsimonious pressures leading to efficiencies in MT (we will call it EMT for efficient MT) is to find the smallest dataset. This may not always be preferable. For example, the most comprehensible dataset to teach a human may not be practically the most efficient. Redundancy and repetitions are sometimes useful tools to reinforce the message for the learner.

Nevertheless, to teach a machine, EMT contributes to efficiency of resources and ensures the machine is taught in the most effective manner. An advantage of EMT is that it reduces demands on the machine\textquoteright s working memory. The definition of EMT is:

\begin{definition}\label{def4}
[\bf{EMT Problem}] Given a model $\hat{\vec{y}} = f(\vec{w} \otimes \vec{x})$ and its parameters $\vec{w}$, find $\hat{\Gamma}$ such that $Err(\hat{\vec{y}},\vec{y})$ and $Cmpx(\hat{\Gamma})$ are minimum.
\end{definition}

$Cmpx(\hat{\Gamma})$ is some complexity metric operating on the dataset required to train the machine; be it simply the number of records/instances, entropy of the sample, or another user defined metric for complexity. A classic regularization term in machine learning reduces the complexity of the model, whereas $Cmpx(\hat{S})$ reduces the complexity of the sample.

The one-off nature of MT/EMT as defined above can be decomposed into a series of chunks that forms a timeseries of incremental MT/EMT (IMT/IEMT) to teach the machine incrementally. 
\begin{definition}
[\bf{IEMT Problem}] Given a model $\hat{\vec{y}} = f(\vec{w} \otimes \vec{x})$ and its parameters $\vec{w}$, find $\hat{\Gamma_t}$ such that $\lim_{\underset{t}{\overset{\infty}{\int}} \hat{\Gamma_t}}$, $Err(\hat{\vec{y}},\vec{y})$ and $Cmpx(\hat{\Gamma})$ are minimum.
\end{definition}

The $\underset{t}{\overset{\infty}{\int}} \hat{\Gamma_t} \rightarrow \hat{\Gamma}$ term is stating that as we accumulate the chunks through integration or summation in discrete cases, in the limit, the cumulative chunks should approach the overall sample $\hat{\Gamma}$ that is needed to approximate $\Gamma$.

Our definition of a curriculum above differs from Bengio et.al.~\cite{bengio2009learning}. In their case, a curriculum is seen as a weighting function on the sample with a higher weight initially given to a simpler problem to learn. Weights then shift to the more difficult problems in a similar manner to the classic Boosting algorithm~\cite{schapire1990strength,freund1995desicion}, where misclassified instances in one iteration receive higher weights in the following iteration, and an ensemble of learners get formed. Our methods, however, will generate modular structures that form a single network, while each module has a `meaning' within an ontological representation.

Curriculum design assumes that $\Gamma$ is known, while IEMT attempts to find ordered sets that in the limit will approximate $\hat{\Gamma}$. Formally, curriculum design partitions $\Gamma$ such that $\underset{o}{\bigcup} \Gamma_o = \Gamma$. In IEMT, each $\hat{\Gamma_t}$ does not need to be a subset of $\Gamma$. These are only guiding lessons to lead the learner towards $\hat{\Gamma}$.

For a linear learner, each $\hat{\Gamma_t}$ results in $\vec{w_t}$. The objective of IEMT is to ensure that as $   \underset{t}{\overset{\infty}{\int}} \hat{\Gamma_t}  \rightarrow \hat{\Gamma}$, thus, from Definition~\ref{def4}, we draw $\lim_{t \rightarrow \infty} \left( \vec{w_t} \right) \rightarrow \vec{w}$. Each $\vec{w_t}$ defines a vector of angles, $\vec{\theta_t}$, with a dimension of $n-1$.

So far, the only condition we have on $\hat{\Gamma_t}$ is that it is as short as possible. A second critical condition we need to establish is that $\hat{\Gamma_t}$ should be interpretable; that is, a meaning is associated with each chunk, and that the sequence of $\hat{\Gamma_t}$ over time is, therefore, explainable by understanding the relationships among the meaning of the individual chunks and their order. This condition is what we call semantic ordering; that is, by sequencing chunks to the machine, with each chunk has a meaning that contributes to the overall logic of the learning process, we will have a lesson plan with identifiable rationale.

To achieve semantic ordering, we will need to design an ontology, $Ont$ for the lessons. Each lesson needs to deliver a concept in that ontology, by generating the required sample of data to learn the concept. As the machine learner learns these lessons and compresses them using its internal representation, the process converges to a machine learner that has learnt the overall task, an explanation of how the learner has learnt the task, and a task-centred interpretation for each module. The above could be captured mathematically as shown in Equations~\ref{Const5}-~\ref{Const6}, where the $Linked$ operator synthesizes the semantic order such that the synthesis covers the overall ontology.

\begin{equation}\label{Const5} Semantic(\hat{\Gamma}_t) \in Ont \end{equation}

\begin{equation}\label{Const6} Linked(\underset{t}{Semantic(\hat{\Gamma}_t)}) = Ont \end{equation}

\section{Case Study}\label{Section4}

In this section, we present a synthetic case study using an artificial neural network to demonstrate scaffolding and the associated semantic representation. The 2D problem represents features calculated on different agents swarming in a space. A swarm is a group of agents that self-synchronise actions to achieve an effect or outcome. A slow swarm means that members of the swarm moves slowly, resulting on a longer time to reach the target; allowing them better opportunities to exchange information and learn from the environment. A fast swarm needs to get to its target fast, with lesser time to learn or exchange information.

\begin{figure}[h]
\centering
\includegraphics[width=0.5\columnwidth]{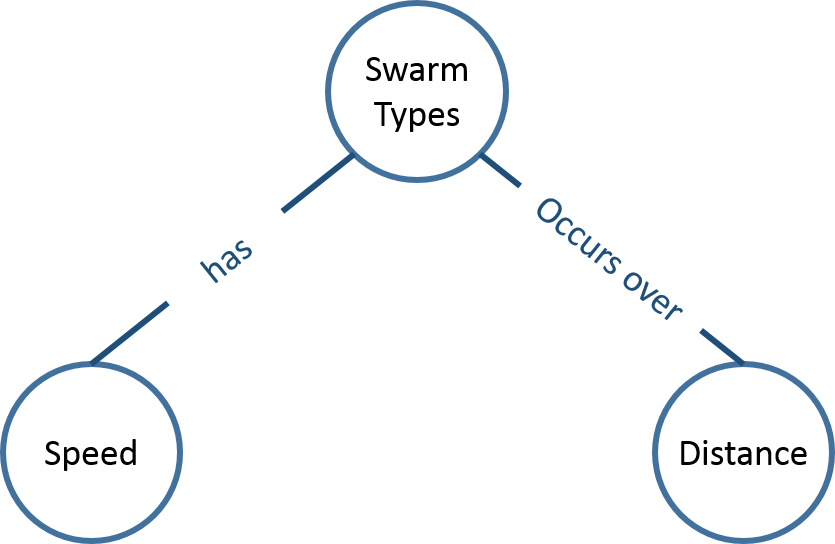}
\caption{Simple Swarm Attack Ontology.}
\label{Ontology} 
\end{figure}

Each cluster is represented by the mode of the time between successive visits of landmarks and the mode of the distance between the landmarks. The simple ontology is represented in Figure~\ref{Ontology}, whereby the primary concept of swarming is associated with two concepts. One is that a swarm has speed and the other is that a swarm occurs spatially over distance.

In this dataset, the ground truth is represented with two inequalities given in Equation~\ref{eq:Eq1}-~\ref{eq:Eq2}. The rationale underpinning them is to define different forms of swarming. Equation~\ref{eq:Eq1} covers the space of slow swarming over medium ranges. Equation~\ref{eq:Eq2} covers the space of fast swarming over longer distances. The intersection of the hyperplanes corresponding to these two inequalities create four polyhedra; three of which are polytopes.

\begin{equation}\label{eq:Eq1} 3 \times Time + 5 \times Distance \ge 150 \end{equation}

\begin{equation}\label{eq:Eq2}  5 \times Time + 2 \times Distance \ge 100 \end{equation}

\begin{figure}[tb]
\centering
\includegraphics[width=0.45\columnwidth]{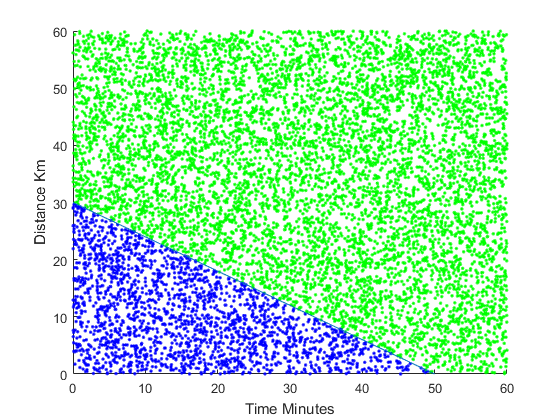}
\includegraphics[width=0.45\columnwidth]{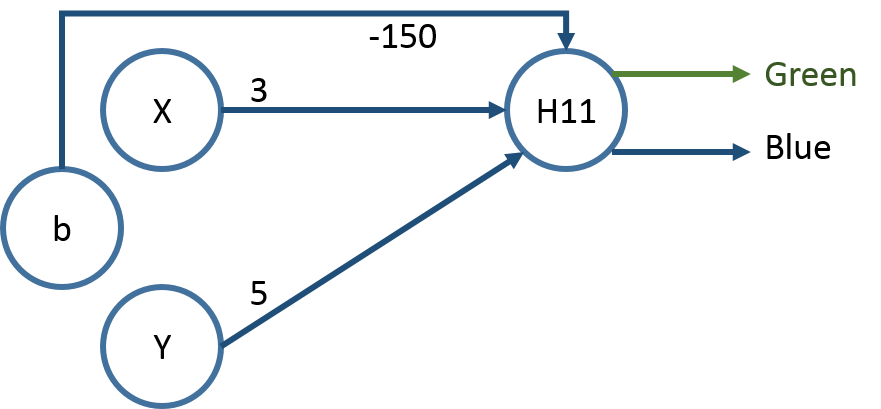}
\includegraphics[width=0.45\columnwidth]{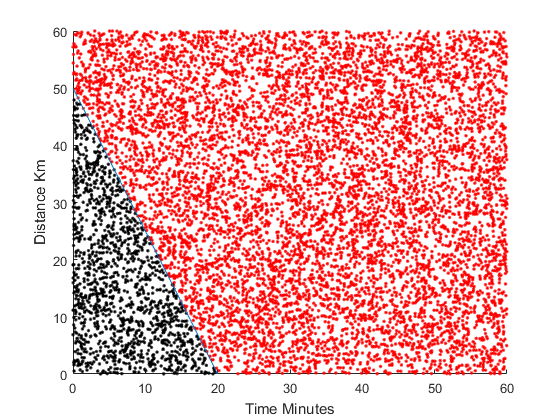}
\includegraphics[width=0.45\columnwidth]{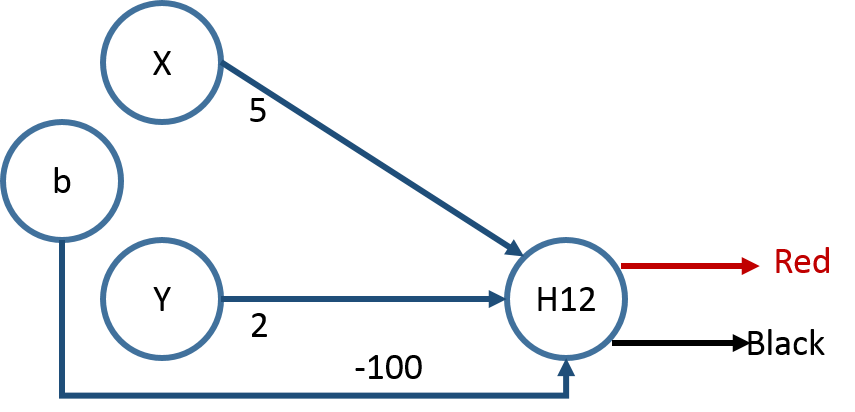}
\caption{Data Chunks.}
\label{DataChunk} 
\end{figure}

If we consider this as the curriculum we wish to deliver to a machine learner, we could use the ontology to guide the sequencing of the datasets to the machine learner. Figure~\ref{DataChunk} presents the samples used to generate the two lessons and the resultant networks. The first two samples were generated to learn the lessons associated with Equations~\ref{eq:Eq1} and ~\ref{eq:Eq2}, respectively;  signifying slow swarming over medium ranges, and fast swarming over longer distances, respectively.

\begin{figure*}[tb]
\centering
\includegraphics[width=0.95\columnwidth]{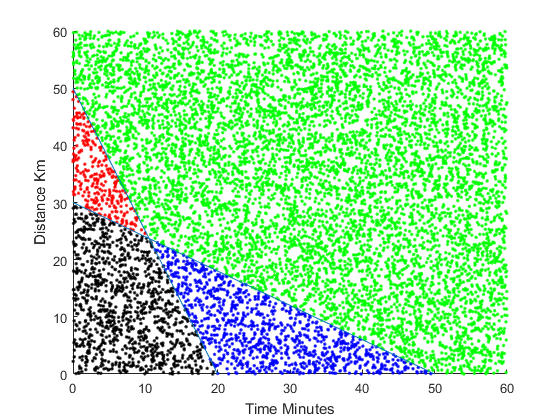}
\includegraphics[width=0.95\columnwidth]{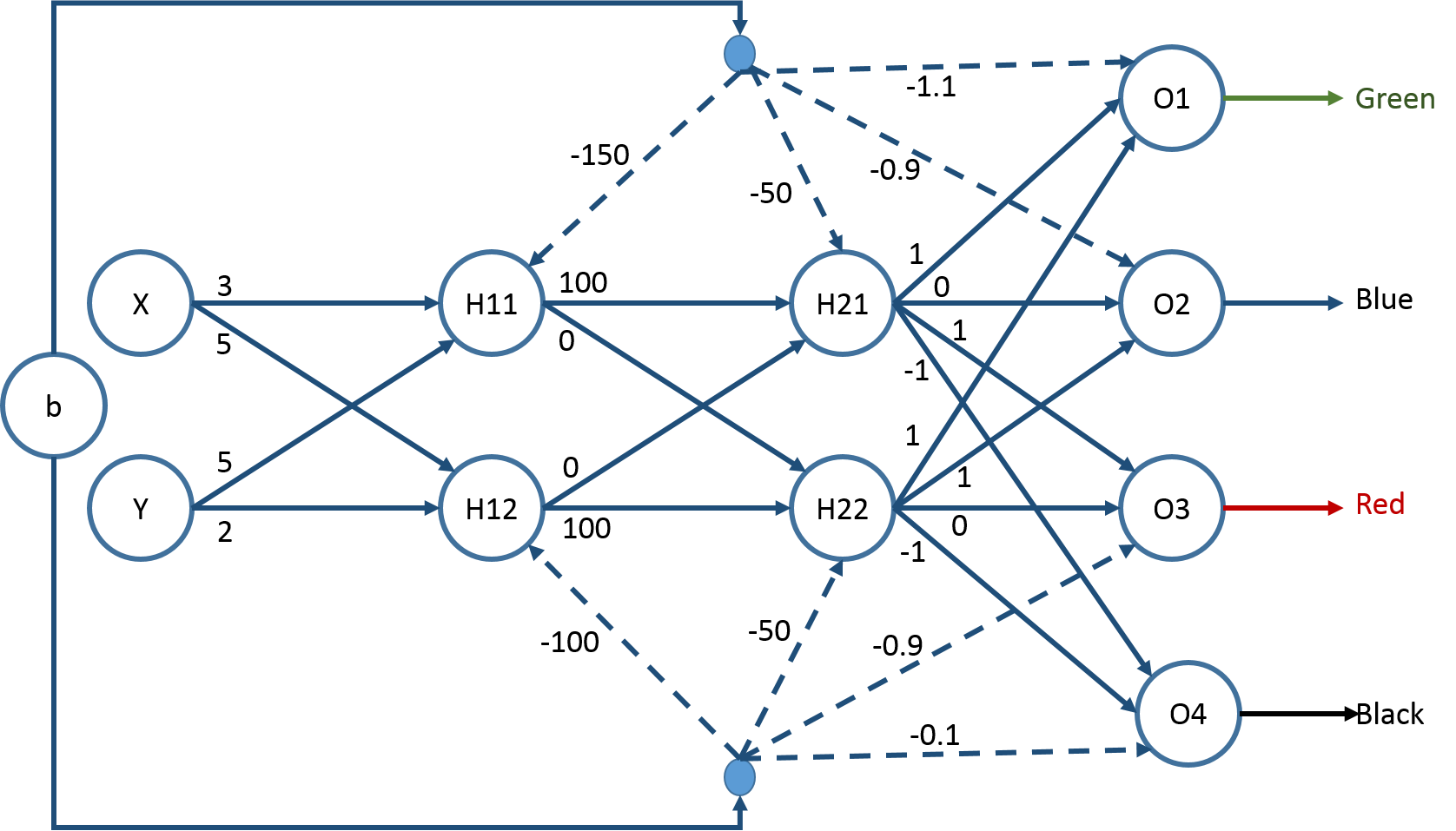}
\caption{Data Chunks.}
\label{Synthesis} 
\end{figure*}

As each concept is learnt by one network, these networks get frozen for the third lesson that synthesizes these concepts to learn the different types of swarming as shown in Figure~\ref{Synthesis}.

\section{Discussion and Conclusion}

Machine teaching is an emerging field with a focus on the experience that gets presented to the machine, or what to learn. It complements machine learning, which focuses on how to learn. The literature is in its infancy, focusing on the data side with less or no emphasis being placed on the pedagogical framework for this educational process. In this paper, we brought concepts from education to introduce machine education, presenting to researchers and scientists the ingredients to educate machines in a similar manner to human education. System dynamics is proposed as an appropriate modelling framework to transform user intent and tasks into the skills expected from the machine and the learning goals. We then presented the concept of scaffolding, which, when coupled with ontology, offers a structured methodology to semantically decompose the problem into data chunks. Together with the system dynamic model, the approach allows the design of a transparent machine learners and offers a methodology that could systematically, experimentally or theoretically verify the learning experience of the machine to ensure its trustworthiness.

\section*{Acknowledgement}
The authors would like to acknowledge the US Office of Naval Research - Global (ONR-G) Grant and the Air Force Office of Scientific Research under award number FA2386-17-1-4054.


\end{document}